\documentclass[conference,a4paper,10pt]{IEEEtran}

\usepackage{graphicx}
\usepackage[table]{xcolor}

\usepackage{amsmath}
\usepackage{amssymb}
\usepackage{amsfonts}
\usepackage{mathtools}
\usepackage{physics}

\usepackage{adjustbox}
\usepackage{multirow}
\usepackage{booktabs}
\usepackage{threeparttable}

\usepackage{hyperref}
\usepackage{cite}

\definecolor{dark}{RGB}{5, 23, 28}
\definecolor{bright}{RGB}{254, 255, 255}
\definecolor{moongrey}{RGB}{214, 218, 221}
\definecolor{guamblue}{RGB}{29, 97, 158}
\definecolor{osakared}{RGB}{236, 97, 159}
\definecolor{abiskogreen}{RGB}{106, 191, 163}
\definecolor{erfoudorange}{RGB}{247, 167, 18}

\newcommand{\subref}[2]{\hyperref[#1]{\ref*{#1}#2}}

\begin{document}

\title{Explaining Anomalies with Tensor Networks}

\author{%
\IEEEauthorblockN{%
Hans Hohenfeld\IEEEauthorrefmark{1}\IEEEauthorrefmark{2},
Marius Beuerle\IEEEauthorrefmark{3}, and
Elie Mounzer\IEEEauthorrefmark{1}
}
\IEEEauthorblockA{\IEEEauthorrefmark{1}%
German Research Center for Artificial Intelligence,
Robotics Innovation Center, 28359 Bremen, Germany \\
Email: hans.hohenfeld@dfki.de, elie.mounzer@dfki.de}
\IEEEauthorblockA{\IEEEauthorrefmark{2}%
QBITFLOW GmbH, 28217 Bremen, Germany}
\IEEEauthorblockA{\IEEEauthorrefmark{3}%
University of Bremen, Robotics Research Group, 28359 Bremen, Germany}
}

\maketitle

\begin{abstract}
    Tensor networks, a class of variational quantum many-body wave functions have attracted
    considerable research interest across many disciplines, including classical machine learning.
    Recently, Aizpurua~et~al.\ demonstrated explainable anomaly detection with matrix product states
    on a discrete-valued cyber-security task, using quantum-inspired methods to gain insight into
    the learned model and detected anomalies. Here, we extend this framework to real-valued data
    domains. We furthermore introduce tree tensor networks for the task of explainable anomaly
    detection.  We demonstrate these methods with three benchmark problems, show adequate predictive
    performance compared to several baseline models and both tensor network architectures'
    ability to explain anomalous samples.
    We thereby extend the application of tensor networks to a broader class of potential problems
    and open a pathway for future extensions to more complex tensor network architectures.
\end{abstract}


\section{Introduction}
\label{sec:intro}
Anomaly detection~\cite{chandola2009, han2022} is the task of identifying patterns or samples in
data that are rare, novel, or deviate from a norm or expectation. It is distinct from, e.g.,
classification, as the available data for a given problem is usually strongly biased toward regular
instances, and samples are (partially) unlabeled. For many applications, such as fraud detection,
medical diagnostics, or monitoring of industrial processes, explaining anomalies, rather than just
identifying them, can benefit informed decision-making by domain experts~\cite{li2023a}.  Tensor
networks~\cite{orus2014a}, originally a technique to express low-rank approximations of otherwise
classically intractable quantum many-body systems, have been proposed for machine learning
applications such as compression of neural networks~\cite{novikov2021, kossaifi2019, nguyen2022,
tomut2024}, supervised~\cite{stoudenmire2016a, stoudenmire2018b, liu2019, sun2020, glasser2020,
reyes2021, wall2021} and unsupervised~\cite{novikov2015, han2018, li2018a, cheng2019, dolgov2020,
hur2023, peng2023a, meiburg2025} learning. Aizpurua et al.\,~\cite{aizpurua2025} recently proposed
an anomaly detection framework based on matrix product states (MPS), demonstrating the identification of
anomalous samples in a cyber-security data set and quantum-inspired techniques to extract rich
information from the trained model to understand the detected anomalies.

Here, we extend this tensor network based anomaly detection framework to real-valued data,
leveraging results on real-valued density estimation with matrix product states~\cite{meiburg2025}.
We furthermore demonstrate the applicability of these techniques to tree tensor networks (TTN) and
compare their predictive performance and ability to explain anomalies. Additionally, we extend the
two-site DMRG-like gradient-based optimization procedure previously introduced for
MPS~\cite{han2018} to tree tensor networks, enabling the adaptation of bond dimensions during training.
We evaluate the tensor networks on three anomaly detection benchmark data sets and compare their
predictive performance to several baseline models before demonstrating their ability to provide
insight into detected anomalies. Our contribution broadens the scope of application for tensor
networks in explainable AI and may foster similar research directions for quantum machine learning
due to the intricate connection between tensor networks and quantum circuits in the latter
field~\cite{huggins2019b, rieser2023a, aizpurua2024a}.

\begin{figure}[t]
    \centering
    \includegraphics{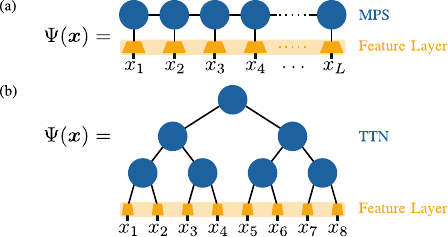}
    \caption{Matrix product state~(a) and tree tensor network~(b). The tensors are contracted with
    an encoded feature vector. For details we refer to Sec.~\ref{sec:method}.}
    \label{fig:tn}
\end{figure}

The remaining paper is structured as follows: In Sec.~\ref{sec:method} we outline the theoretical
background, learning methods, explainability metrics, and experimental setup. Afterward, we present
and discuss our findings in Sec.~\ref{sec:results}, before outlining limitations and possible future
research directions in Sec.~\ref{sec:conclusion}.

\begin{figure*}[t]
    \centering
    \includegraphics{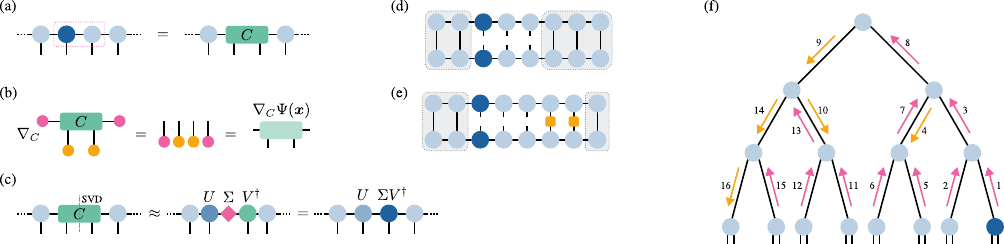}
    \caption{In each training step, the canonical center (dark blue) is contracted with its neighbor in the
        traversal direction~(a). The gradient $\nabla_{C}\Psi(\boldsymbol{x})$ is computed by
        contracting the network around $C$ and removing $C$. The outer product of the remaining
        vectors is the gradient tensor~(b). After each update step, the contracted tensor undergoes a
        truncated singular value decomposition. The singular values are contracted in the direction
        of traversal to move the canonical center~(c). To compute the reduced density matrix of a
        sub-system, the tensor network is contracted with its transpose, leaving the physical bonds
        of the sites of the sub-system open~(d). To compute the reduced density matrix for a
        conditional marginal distribution, the sites the distribution is conditioned on are fixed to
        $\ketbra{x_{i}=a_{i}}, a_{i}\in\mathcal{I}$ in the contraction~(e). The areas with gray
        backgrounds contract to the identity. The traversal order for the two-site update on TTNs
        from the canonical center (right-most leaf) is akin to a depth-first traversal with back-tracking,
        where all but the last leaf are only traversed upwards~(f). All computations in (a)-(e) are
        shown for matrix product states but work for TTNs analogously.}
    \label{fig:algos}
\end{figure*}

\section{Method}
\label{sec:method}

\subsection{Tensor Networks}
\label{ssec:tn}

The state vector of a quantum many-body system of size $L$ with local $d$-dimensional state space
$\{\sigma_{i}\}$ is given by
\begin{equation}
    \ket{\psi} \coloneq \sum_{\sigma_{1},\ldots,\sigma_{L}}
        c_{\sigma_{1},\ldots,\sigma_{L}}\ket{\sigma_{1},\ldots,\sigma_{L}}.
\end{equation}
This general and exact representation does not take known properties of many physical systems of
interest, such as locality, correlation length, or criticality, into account~\cite{orus2014a}.
Writing the $d^{L}$ coefficients $c_{\sigma_{1},\ldots,\sigma_{L}}\in\mathbb{C}$ into a tensor
$\Psi(\boldsymbol{\sigma})=c_{\boldsymbol{\sigma}}$ with
$\boldsymbol{\sigma}=(\sigma_{1},\ldots,\sigma_{L})$ enables a compressed re-parameterization of
$\ket{\psi}$ as a tensor network. The first class of tensor network states we consider in this work
are \textit{matrix product states} (MPS), decomposing $\Psi(\boldsymbol{\sigma})$
into~\cite{schollwock2011}
\begin{equation}
    \Psi(\sigma_{1},\ldots,\sigma_{L}) \coloneq
        \smashoperator{\sum_{\{\alpha\}}}
        A_{\alpha_{1}}^{\sigma_{1}}
        A_{\alpha_{1}\alpha_{2}}^{\sigma_{2}}
        \cdots
        A_{\alpha_{L-1}}^{\sigma_{L}}.
\end{equation}
Each $A^{\sigma_{i}}$ is a $D_{i-1}\times D_{i}$ matrix with $D_{0}=D_{L}=1$, graphically
illustrated in Fig.~\subref{fig:tn}{a}. An exact decomposition leads to $D_{i}\leq d^{\frac{L}{2}}$,
for many applications, such as the ground state computation of local Hamiltonians with gapped
eigenspectrum on a 1-dimensional lattice, a rank-reduced approximation suffices, as those states
follow an area law of entanglement entropy~\cite{hastings2004}. For any bi-partition of $\ket{\psi}$
into sub-systems $\mathcal{X}$ and $\mathcal{Y}$ at bond $j$, the entanglement entropy $S$ is bound
by $S(\rho_{\mathcal{X}})\leq \log D_{j}$, enabling faithful representation of area-law states with
$D_{i}\ll d^{\frac{L}{2}}$~\cite{verstraete2006}, typically $D_{i}\sim\mathcal{O}(100)$.

\textit{Tree tensor networks} (TTN) were introduced to study strongly correlated systems with
non-local interactions~\cite{murg2010}. For a balanced binary tree, as shown in
Fig.~\subref{fig:tn}{b}, $\Psi(\boldsymbol{\sigma})$ is given by
\begin{equation}
    \Psi(\boldsymbol{\sigma})
        \coloneq \sum_{\{\alpha\}} A_{\alpha_{1}\alpha_{2}}
        \smashoperator{\prod_{n=1}^{L/2 - 2}} A_{\alpha_{n}\alpha_{2n+1}\alpha_{2n+2}}
        \smashoperator{\prod_{n=1}^{L/2}} A_{\alpha_{n+L/2-2}}^{\sigma_{2n-1}\sigma_{2n}}.
\end{equation}
The binary tree construction is no necessity. TTNs can be expressed by arbitrary trees and their
structure even be subject to optimization~\cite{hikihara2023a}.

\subsection{Unsupervised Learning with Tensor Networks}
\label{ssec:unsupervised}

A tensor network state can be interpreted as a probability distribution of $d^{L}$ basis states. Let
$\boldsymbol{x}\in\{0,1\}^{L}$ and $d=2$ then
\begin{equation}
    P(\boldsymbol{x}) = |\braket{\psi}{\boldsymbol{x}}|^{2} = |\Psi(\boldsymbol{x})|^{2},
\end{equation}
with $\boldsymbol{x}\coloneq \ket{x_{1}}\otimes\ket{x_{2}}\otimes\cdots\otimes\ket{x_{L}}$. For a
data set $\mathcal{D}\subseteq\{0,1\}^{L}$, we can optimize the tensor network to represent the
probability distribution over $\mathcal{D}$ by minimizing the negative log-likelihood
loss~\cite{han2018, cheng2019}
\begin{equation}
    \mathcal{L}\coloneq -\frac{1}{|\mathcal{D}|}\sum_{\boldsymbol{x}\in\mathcal{D}}
        \log P(\boldsymbol{x}).
\end{equation}
To train the tensor network, we start with one site as canonical center~\cite{schollwock2011} and
traverse the network. In each training step, the current canonical center is contracted with its
neighboring site in direction of traversal, as shown in Fig.~\subref{fig:algos}{a}. The contracted
site $C$ is then updated with one or several steps of (stochastic) gradient descent with the
gradient
\begin{equation}
    \nabla_{C}\mathcal{L}= -\frac{2}{|\mathcal{D}|}\sum_{\boldsymbol{x}\in\mathcal{D}}
        \frac{\nabla_{C}\Psi(\boldsymbol{x})}{\Psi(\boldsymbol{x})},
\end{equation}
updating $C$ by $C \leftarrow C - \eta \nabla_{C}\mathcal{L}$,
with a learning rate $0<\eta\ll 1$. The procedure to compute $\nabla_{C}\Psi(\boldsymbol{x})$
is shown if Fig.~\subref{fig:algos}{b}. After updating, $C$ is decomposed using a truncated SVD,
where we control the bond dimension by discarding singular values below a threshold:
\begin{equation}
    C\approx U\Sigma V^{\dagger}.
\end{equation}
As shown in Fig.~\subref{fig:algos}{c}, the singular values are then contracted into the next site
in the direction of traversal, making it the new canonical center. For matrix product states, the
traversal is a sweeping motion from left to right and back. Tree tensor networks have been trained
with single site updates in previous works~\cite{cheng2019}, here we adapt the two-site learning
procedure to a depth-first traversal with backtracking, illustrated in Fig.~\subref{fig:algos}{f}.

\subsection{Encoding Real-Valued Data}
\label{ssec:encoding}

To encode real-valued data into a feature layer as shown in Figure~\ref{fig:tn}, we leverage the
framework introduced in Ref.~\cite{meiburg2025}. Given a sample $\boldsymbol{x}\in\mathcal{D}$ with
$\boldsymbol{x}=(x_{1},\ldots,x_{L})$, each feature $x_{i}\in\mathcal{I}$ is encoded with an
isometric feature map $\xi:\mathcal{I}\to\mathbb{R}^{N}$ composed of $N$ feature functions
\begin{equation}
    \xi(x_{i})\coloneq \ket{x_{i}} = \begin{bmatrix}
        g_{1}(x_{i}) \\ g_{2}(x_{i}) \\ \vdots \\ g_{N}(x_{i})
    \end{bmatrix}.
\end{equation}
The authors of Ref.~\cite{meiburg2025} discuss several feature functions and their induced priors.
We found the best results by rescaling all features to $\mathcal{I}=[0,1]$ and encode with
\begin{equation}
    g_{n}(x_{i})\coloneq \sqrt{2n+1}\tilde{P}_{n}(x_{i}).
\end{equation}
$\tilde{P}_{n}:\mathbb{R}\to\mathbb{R}$ is the $n$-th shifted Legendre polynomial
\begin{equation}
    \label{eq:ortho}
    \tilde{P}_{n}(x_{i}) \coloneq \frac{1}{n!}\dv[n]{x_{i}}(x_{i}^{2}-x_{i})^{n},
\end{equation}
fulfilling the orthogonality relation
\begin{equation}
    \int_{0}^{1}\tilde{P}_{m}(x)\tilde{P}_{n}(x)\dd x = \frac{1}{2n+1}\delta_{mn}.
\end{equation}
The factor $\sqrt{2n+1}$ scales the polynomials to an orthonormal basis.

\subsection{Extracting Information}
\label{ssec:extraction}

Tensor networks are a representation of a class of quantum states. Thus, we can apply
quantum information theoretic methods to analyze the trained model and data samples of
interest. In Ref.~\cite{aizpurua2025}, the authors established the reduced density matrix
corresponding to a subset of features as the primary tool for this purpose. Conceptually, the reduced
density matrix corresponds to the marginal distribution the model assigns to the sub-system.
Fig.~\subref{fig:algos}{d} shows the contraction of an MPS with its conjugate transpose, leaving the
physical indices of the sub-system open, which, after re-shaping, computes the reduced density matrix.
Here, we operate on real-valued data, in principle, requiring integrating over the variables being
marginalized out. Given the isometric encoding map, we have
\begin{equation}
    \int_{\mathcal{I}}\ketbra{x_{i}} = \mathbb{I},
\end{equation}
and this step can be omitted. The conditional reduced density matrix, corresponding to the
conditional marginal distribution of a sub-systems of features, is computed by the contraction shown
in Fig.~\subref{fig:algos}{e}. The variables to condition the distribution on are fixed by
introducing $\ketbra{x_{i}=a_{i}}, a_{i}\in\mathcal{I}$ into the contraction.
If we make one of the sites within the boundaries of the sub-system the canonical center before either of
these contractions, the areas shaded in gray in Fig.~\subref{fig:algos}{d} and~\subref{fig:algos}{e}
contract to the identity~\cite{schollwock2011}. When working with discrete-valued data encoded into
basis states, the multinomial distribution over the features of the sub-system can be directly read
from the diagonal of the reduced density matrix.
This information is not readily available for real-valued data encoded in an isometric feature layer,
but the reduced density matrix induces a quasi-probability density function
\begin{equation}
    \label{eq:density}
    q(\boldsymbol{x})\coloneq \trace\left( \ketbra{\boldsymbol{x}}\rho_{\mathcal{X}} \right),
\end{equation}
which after normalization by its integral can be used to compute expected value, expected
variance, and the expected covariance matrix for the sub-system $\mathcal{X}$.
From the reduced density matrix two further measures of interests can be computed.
That is, the von Neumann entropy
\begin{equation}
    S(\rho_{\mathcal{X}}) \coloneq -\trace(\rho_{\mathcal{X}}\log\rho_{\mathcal{X}}),
\end{equation}
and the mutual information $I$ of two sub-systems $\mathcal{X}$ and $\mathcal{Y}$
\begin{equation}
    \label{eq:mutual}
    I(\mathcal{X};\mathcal{Y}) \coloneq S(\rho_{\mathcal{Y}}) + S(\rho_{\mathcal{Y}})
        - S(\rho_{\mathcal{X},\mathcal{Y}}).
\end{equation}
With these, we can analyze the learned correlations and dependencies between individual and groups
of features.

\subsection{Learning Setup}
\label{ssec:setup}

We use three different benchmark data sets for this study. Their main properties are summarized in
Table~\ref{tab:datasets}. The \texttt{ECG5000}~\cite{chen2015} data set contains about 5,000
heartbeats from an ECG recording of a patient with congestive heart failure.  Each heartbeat is
normalized in time to equal length and described by 140 features.
\texttt{Satellite}~\cite{srinivasan1993} contains 6,435 multi-spectral satellite images, each sample
being a 3$\times$3 patch recorded in four different bands. The third set,
\texttt{Spambase}~\cite{hopkins1999}, is a collection of 4,601 email samples, described by the word
count of 57 different key characters or words used to identify spam emails.  $N_{\text{MPS}}$ and
$N_{\text{TTN}}$ in Table~\ref{tab:datasets} refer to the number of encoding feature functions for
each data set and tensor network type. The selected data sets provide a variety of
application domains and are commonly used in the anomaly detection literature. Thus, baseline
performance comparisons are readily available. Yet, while being instructive for the techniques
discussed here, they are relatively small in scale. This choice is a consequence of the considerable
overhead in computational resources required for the tensor network models, which we discuss in
Sec.~\ref{ssec:complexity}.

All models are trained unsupervised. To this end, we prepare each data set to contain 95\% regular
samples and 5\% anomalies, either unlabeled. The anomalies are composed in equal parts of those
provided with the data set and generated anomalies, for which we use the techniques described in
Ref.~\cite{han2022}. The prepared data sets are then split into a stratified
10-fold. Each model is trained and evaluated ten times, with one of the folds as a hold-out set. We
perform two evaluation tasks. First, we assess the model's ability to separate regular from
anomalous samples in the nine training folds. Afterward, we conduct an inductive
evaluation with the hold-out set.

\begin{table}[t]
    \begin{threeparttable}
        \caption{Properties of the data sets used in this study.}
        \label{tab:datasets}
        \begin{tabular}{ccccccc}
            \toprule
            \textbf{Data set} & \textbf{Ref.} & $\boldsymbol{L}$\tnote{a} &
            \textbf{Samples} & \textbf{Anomalies} &
            $\boldsymbol{N_{\text{MPS}}}$\tnote{b} & $\boldsymbol{N_{\text{TTN}}}$\tnote{c} \\
            \cmidrule(lr{.5em}){1-1}
            \cmidrule(lr{.5em}){2-2}
            \cmidrule(lr{.5em}){3-3}
            \cmidrule(lr{.5em}){4-4}
            \cmidrule(lr{.5em}){5-5}
            \cmidrule(lr{.5em}){6-6}
            \cmidrule(lr{.5em}){7-7}
            \texttt{ECG5000} & \cite{chen2015} & 140 & 4,998 & 2,079 & 4 & 4 \\
            \texttt{Satellite} & \cite{srinivasan1993} & 36 & 6,435 & 2,036 & 5 & 5 \\
            \texttt{Spambase} &\cite{hopkins1999} & 57 & 4,601 & 1,813 & 6 & 5 \\
            \bottomrule
        \end{tabular}
        \begin{tablenotes}
            \item[a] Number of features
            \item[b] Number of encoding feature functions for MPS
            \item[c] Number of encoding feature functions for TTN
        \end{tablenotes}
    \end{threeparttable}
\end{table}

To assess the anomaly detection capability of both tensor networks, before demonstrating their
ability to explain anomalous samples, we compare their performance to three baseline models.  We use
a one-class support vector machine (SVM)~\cite{scholkopf2001}, an isolation forest
(IFO)~\cite{liu2008} and an autoencoder (AE)~\cite{hinton2006a} with two hidden layers in the
encoder and decoder part.  Before experiments, the hyper-parameters of all models and for all data
sets were tuned extensively with Optuna~\cite{akiba2019a}.  We implemented the experimental setup
using Jax~\cite{bradbury2018a} and opt\_einsum~\cite{smith2018}, code is available under an
open-source license at Ref.~\cite{hohenfeld2025}.

\section{Results}
\label{sec:results}

\subsection{Anomaly Detection Performance}
\label{ssec:anomaly}

In Table~\ref{tab:auc}, we report the mean area under the curve for the receiver operating
characteristic (AUCROC) and its standard deviation for all models, data sets over the ten
repetitions of the separation and inductive anomaly detection tasks. In general, all models perform
adequately and in line with previous results reported on these or similar data sets in the
literature~\cite{han2022} for unsupervised methods. Notably, the mean performance remains almost
unchanged, going from the separation task to the evaluation on the hold-out set, with only the
standard deviation increasing slightly. Setting the threshold for flagging a sample anomalous is a
trade-off between sensitivity and specificity, which regularly is domain and task-dependent. Since our
primary focus here is explainability and not task-specific predictive performance, we set the
threshold to the intersection between true-positive and true-negative rates, which provided sufficient
results for this study.

\begin{table}[t]
    \centering
    \caption{Mean AUCROC score and std.\ deviation on both tasks.}
    \label{tab:auc}
    \begin{tabular}{cccccccc}
        \toprule
        & &
        \multicolumn{2}{c}{\textbf{ECG5000}} &
        \multicolumn{2}{c}{\textbf{Satellite}} &
        \multicolumn{2}{c}{\textbf{Spambase}} \\
        \cmidrule(lr{.5em}){3-4}
        \cmidrule(lr{.5em}){5-6}
        \cmidrule(lr{.5em}){7-8}
        \textbf{Task} & \textbf{Model} &
        \textbf{Mean} & \textbf{Std.} &
        \textbf{Mean} & \textbf{Std.} &
        \textbf{Mean} & \textbf{Std.} \\
        \cmidrule(lr{.5em}){1-1}
        \cmidrule(lr{.5em}){2-2}
        \cmidrule(lr{.5em}){3-3}
        \cmidrule(lr{.5em}){4-4}
        \cmidrule(lr{.5em}){5-5}
        \cmidrule(lr{.5em}){6-6}
        \cmidrule(lr{.5em}){7-7}
        \cmidrule(lr{.5em}){8-8}
        \parbox[t]{1mm}{\multirow{5}{*}{\rotatebox[origin=c]{90}{\textbf{Separation}}}}
        & \textbf{MPS} & 0.92 & 0.02 & \cellcolor{abiskogreen!50}\textbf{0.90} & 0.01 & 0.83 & 0.01 \\
        & \textbf{TTN} & \cellcolor{abiskogreen!50}\textbf{0.96} & 0.00 & 0.89 & 0.01 & 0.82 & 0.01 \\
        & \textbf{SVM} & 0.95 & 0.00 & 0.80 & 0.01 & 0.77 & 0.01 \\
        & \textbf{IFO} & 0.93 & 0.00 & \cellcolor{abiskogreen!50}\textbf{0.90} & 0.01 &
            \cellcolor{abiskogreen!50}\textbf{0.88} & 0.01 \\
        & \textbf{AE}  & 0.93 & 0.01 & 0.87 & 0.03 & 0.81 & 0.01 \\
        \midrule
        \parbox[t]{1mm}{\multirow{5}{*}{\rotatebox[origin=c]{90}{\textbf{Inductive}}}}
        & \textbf{MPS} & 0.92 & 0.03 & 0.89 & 0.05 & 0.83 & 0.06 \\
        & \textbf{TTN} & \cellcolor{abiskogreen!50}\textbf{0.96} & 0.01 & 0.89 & 0.05 & 0.82 & 0.05 \\
        & \textbf{SVM} & 0.95 & 0.02 & 0.80 & 0.07 & 0.77 & 0.06 \\
        & \textbf{IFO} & 0.93 & 0.02 & \cellcolor{abiskogreen!50}\textbf{0.90} & 0.05
            & \cellcolor{abiskogreen!50}\textbf{0.88} & 0.06 \\
        & \textbf{AE}  & 0.94 & 0.03 & 0.86 & 0.06 & 0.81 & 0.05 \\
        \bottomrule
    \end{tabular}
    \begin{tablenotes}
        \item Mean and standard deviation for the area under the ROC curve.
        \item Best result per task/data set marked with green background.
    \end{tablenotes}
\end{table}

\begin{figure*}[t]
    \centering
    \includegraphics{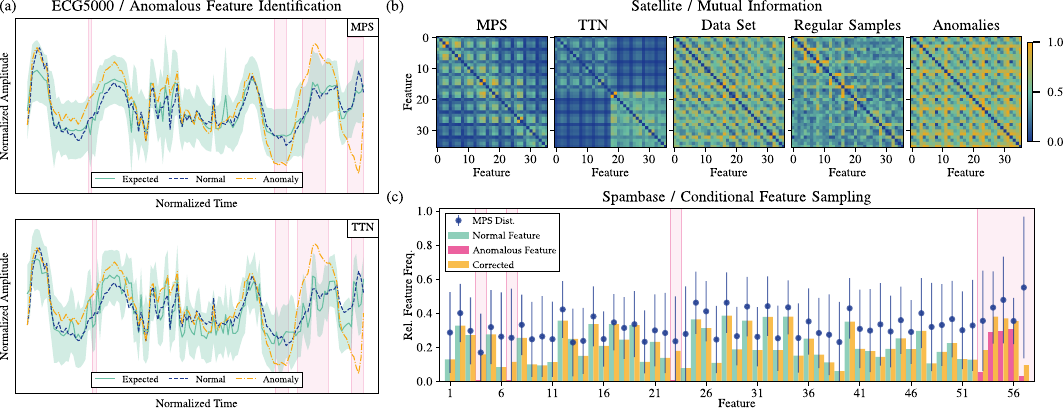}
    \caption{Three examples of information extraction from the trained tensor network models. Both,
    MPS and TTN correctly identified an anomalous heartbeat in the \texttt{ECG5000} data set~(a).
    Beyond the detection, both models can identify the parts of the ECG, that explain the anomaly,
    in this example the sharp spike and reversal of amplitude at the end. This information can
    support a domain expert to judge the presented data. Notably, the TTN learned a sharper and
    narrower distribution, than the MPS. The all-to-all feature mutual information extracted from
    the MPS and TTN~(b) shows a structural pattern, that can also be seen in the normal
    samples from the \texttt{Satellite} data set, but not in the polluted data, the models were
    trained on. For better visibility, values are scaled to the interval $[0, 1]$ and the diagonal
    set to 0. The tensor networks not only enable identifying anomalous features of a sample~(c)
    but also computing their expected values, conditioned on the normal features of the sample,
    providing additional insight into a detected anomaly.}
    \label{fig:plots}
\end{figure*}

\subsection{Identifying Anomalous Features}
\label{ssec:features}

After detecting anomalous samples, understanding what features of the sample are outside a defined
norm benefits diagnostics and informs decision-making by domain experts. This information is readily
available from the tensor network models with the techniques discussed in
Sec.~\ref{ssec:extraction}. As an example, we show the per-feature distribution learned by a matrix
product state and by a tree tensor network on the \texttt{ECG5000} data set in
Fig.~\subref{fig:plots}{a}. After normalizing the quasi-probability density defined in
Eq.~\eqref{eq:density} for each feature, we compute its expected value and expected standard
deviation, depicted by the green line and the area around it in Fig.~\subref{fig:plots}{a}.
Additionally, a normal sample (blue) and ECG recording flagged as anomalous (yellow) are shown. The
anomalous sample has several ranges of features outside the expected standard deviation, indicated
by the red-shaded areas. Here, the sharp rise and following inversion of the amplitude at the end of
the sample stand out, indicating a type of arrhythmia known as premature ventricular
contraction~\cite{gerstenfeld2019}. Setting the appropriate range relative to the per-feature
distribution to consider a feature anomalous is a decision to be made by domain experts on a
task-dependent basis. We additionally find that the TTN learned a narrower distribution with a
sharper distinction of features. Consequently, the anomalous features stand out more pronounced
compared to the model learned by the MPS.

\subsection{Inspecting Model Structure}
\label{ssec:mutual}

The trained tensor networks also provide insight into the structure captured by the model.
Fig.~\subref{fig:plots}{b} shows the all-to-all feature mutual information computed from an MPS and
a TTN trained on the \texttt{Satellite} data set. Next to it, the same measure of feature
dependencies is shown for the entire polluted data set, as well as separately for the regular and
anomalous samples. We determined the latter three with a low-bias histogram-based
estimator~\cite{hacine-gharbi2012}. All values have been scaled to the interval $[0,1]$ and the
diagonal set to $0$, for better visibility. Both tensor networks reveal a similar, structured
pattern that corresponds to the structure of the data set discussed in Sec.~\ref{ssec:setup}. The
same pattern can be recognized, albeit less sharp, in the mutual information computed for the
regular samples, but not for the polluted data set, we used to train the models. For a real-world
task, such details may not be known upfront, and the trained tensor network model can support
analyzing and understanding structure and feature dependence in the data, separating regular from
anomalous samples during training.

\subsection{Conditional Sampling}
\label{ssec:conditional}

For some applications, it can be helpful to understand how anomalous features should be
different, given the context of the remaining features of a sample. Fig.~\subref{fig:plots}{c} shows
such an evaluation for the \texttt{Spambase} data set. Several features, which are individual
characters or words present in an email, are flagged, as their relative frequency is outside or at
the margins of the per-feature expected standard deviation computed from a trained matrix product
state. As described in Sec.~\ref{ssec:extraction}, we can construct a conditional reduced density
matrix, fixing the features considered normal, and use it to compute the expected value of the
resulting conditional marginal distribution. We note that feature 57, after
computing its expected value from the conditional marginal distribution, is still outside the
expected standard deviation of the same feature's marginal distribution, emphasizing that neither
of the presented techniques should be considered in isolation nor that one is preferable over the
other. Their applicability and usefulness rather has to be assessed in combination and in the context
of the task, to provide the required understanding of detected anomalies.

\subsection{Model Complexity \& Numerical Precision}
\label{ssec:complexity}

Both tensor network models provide valuable insight to understand anomalous samples as well as
the structure of the data set used for training. This additional utility, does not come without
a cost. The tensor network models are notably larger compared to the autoencoder we used as a
baseline model, with between 1.5 and 2.5 times the trainable parameters. Combined with the
inherently sequential learning algorithms, optimizing two sites at a time, make their training
computationally demanding and potentially time and resource prohibitive for high-dimensional
data sets. Another aspect contributing to this limitation is numeric stability and precision. A
tensor network is an approximation of a quantum state, assigning a probability to each state encoded
in the feature layer described in Sec.~\ref{ssec:features}. As a consequence, for each given basis
$\ket{\boldsymbol{\sigma}}$, the total probability has to be distributed among $d^{L}$ basis states,
which for sufficiently large $d$ and $L$ regularly leads to small probabilities during
initialization and training, requiring double-precision floating point parameters for the
relatively small data sets considered in this study.

\section{Conclusion}
\label{sec:conclusion}

Here, we extended the framework for explainable anomaly detection with matrix product
states, introduced in Ref.~\cite{aizpurua2025} to real-valued data and
tree tensor networks. We showed that both tensor network architectures perform adequately in
unsupervised anomaly detection compared to several baseline models on three different benchmark data
sets. Furthermore, we demonstrated how quantum-inspired techniques can be used in the
real-valued domain, to extract information about anomalous features, gain insight into the learned
data set structure and feature dependencies, as well the computation of expected values for individual
features conditioned on parts of a sample. Thereby, we extend the scope of applicability for
explainable anomaly detection with tensor networks to a broader class of potential problems. Given
their roots in the study of quantum many-body systems and their connection to quantum machine
learning~\cite{huggins2019b, rieser2023a, aizpurua2024a}, tensor network based techniques
in machine learning learning can bridge the gap between the current small-scale quantum machine
learning in the NISQ era and potential future learning methods on large-scale, fault-tolerant
quantum computers. One avenue of future research to this end could be methods to compute measures for
explainable machine learning as presented here, directly from a quantum circuit based tensor network
model as well as the efficient transferability of classically (pre-)trained tensor networks onto
quantum circuits. A further natural extension of our work would be the evaluation of the presented
techniques to other types of tensor networks such as PEPS~\cite{verstraete2004} and
MERA~\cite{vidal2008} and in-depth comparison of their capabilities in a wider variety of
anomaly detection tasks. Furthermore, we intend to address the computational constraints in terms of
model size and training time discussed in Sec.~\ref{ssec:complexity} in future work.

\section*{Acknowledgment}
\addcontentsline{toc}{section}{Acknowledgment}

This work was funded by the German Ministry of Economic Affairs and Climate Action (BMWK) and the
German Aerospace Center (DLR) in project QuDA-KI (grant no.\ 50RA2206).  We thank Borja Aizpurua for
the helpful discussion and U\u{g}ur Bolat, Lukas Groß, and Gunnar Schönhoff for their feedback on
the manuscript.

\clearpage


\bibliographystyle{IEEEtran}
\bibliography{IEEEabrv, main}

\end{document}